\documentclass{article}

\usepackage[accepted]{include/arxiv}

\usepackage{times}
\usepackage{graphicx} 

\usepackage{natbib}

\usepackage{algorithm}
\usepackage{algorithmicx}
\usepackage[noend]{algpseudocode}
\usepackage{changepage}

\algrenewcommand\algorithmicindent{1.0em}
\renewcommand{\algorithmiccomment}[1]{\hskip1em$\triangleright$ #1}

\usepackage{hyperref}

\usepackage{graphicx}
\usepackage{comment}
\usepackage{amsmath}
\usepackage{amssymb}
\usepackage{siunitx}
\usepackage{wrapfig}
\usepackage{todonotes}
\usepackage{enumitem}
\usepackage{siunitx}
\usepackage{dsfont}
\usepackage{titling}

\newcommand{\E}{\mathbb{E}}

\newcommand{\be}{\begin{equation}}
\newcommand{\ee}{\end{equation}}
\newcommand{\bea}{\begin{eqnarray}}
\newcommand{\eea}{\end{eqnarray}}
\newcommand{\beaa}{\begin{eqnarray*}}
\newcommand{\eeaa}{\end{eqnarray*}}

\DeclareMathAlphabet{\mathpzc}{OT1}{pzc}{m}{n}

\DeclareMathOperator*{\argmax}{arg\,max}

\newcommand{\MLP}[1]{{\ensuremath{\text{MLP}(#1)}}}

\newcommand{\figspace}{\vskip -0.5em}

\icmltitlerunning{Learning to Communicate to Solve Riddles with
	Deep Distributed Recurrent Q-Networks}

\begin{document}

\twocolumn[
\icmltitle{Learning to Communicate to Solve Riddles \\ with
	Deep Distributed Recurrent Q-Networks}

\icmlauthor{Jakob N. Foerster$^{1,\dagger}$}{jakob.foerster@cs.ox.ac.uk}
\icmlauthor{Yannis M. Assael$^{1,\dagger}$}{yannis.assael@cs.ox.ac.uk}
\icmlauthor{Nando de Freitas$^{1,2,3}$}{nandodefreitas@google.com}
\icmlauthor{Shimon Whiteson$^1$}{shimon.whiteson@cs.ox.ac.uk}
\icmladdress{$^1$University of Oxford, United Kingdom\\ $^2$Canadian Institute for Advanced Research, CIFAR NCAP Program \\ $^3$Google DeepMind}

\icmlkeywords{deep reinforcement learning, machine learning, ICML, distributed, multi-agent}

\vskip 0.15in
]

\begin{abstract}
We propose \emph{deep distributed recurrent Q-networks} (DDRQN), which enable teams of agents to learn to solve communication-based coordination tasks.  In these tasks, the agents are not given any pre-designed communication protocol.  Therefore, in order to successfully communicate, they must first automatically develop and agree upon their own communication protocol.  We present empirical results on two multi-agent learning problems based on well-known riddles, demonstrating that DDRQN can successfully solve such tasks and discover elegant communication protocols to do so.  To our knowledge, this is the first time deep reinforcement learning has succeeded in learning communication protocols. In addition, we present ablation experiments that confirm that each of the main components of the DDRQN architecture are critical to its success.

\end{abstract}

\section{Introduction}
\label{sec:introduction}
In recent years, advances in deep learning have been instrumental in solving a number of challenging reinforcement learning (RL) problems, including high-dimensional robot control  \cite{levine2015end,assael2015data,Watter:2015}, visual attention \cite{Ba:2015}, and the \emph{Atari learning environment} (ALE)~\cite{Guo:2014,Mnih:2015,stadie2015incentivizing,Wang:2015duel,Schaul:2015,van2015deep,oh2015action,Bellemare2015Persistent,Nair:2015}.

The above-mentioned problems all involve only a single learning agent.  However, recent work has begun to address multi-agent deep RL.  In competitive settings, deep learning for Go \cite{Maddison:2015,silver2016mastering} has recently shown success.  In cooperative settings, \citet{tampuu2015multiagent} have adapted \emph{deep Q-networks} \cite{Mnih:2015} to allow two agents to tackle a multi-agent extension to ALE.  Their approach is based on \emph{independent Q-learning} \cite{Shoham:2007,MASfoundations09,Zawadzki:2014}, in which all agents learn their own $Q$-functions independently in parallel.

However, these approaches all assume that each agent can fully observe the state of the environment.  While DQN has also been extended to address partial observability \cite{hausknecht2015deep}, only single-agent settings have been considered.  To our knowledge, no work on deep reinforcement learning has yet considered settings that are both partially observable and multi-agent. 

Such problems are both challenging and important. In the cooperative case, multiple agents must coordinate their behaviour so as to maximise their common payoff while facing uncertainty, not only about the hidden state of the environment but about what their teammates have observed and thus how they will act.  Such problems arise naturally in a variety of settings, such as multi-robot systems and sensor networks \cite{Mataric199773,Fox2000325,Gerkey2004939,OlfatiSaber2007215,Cao2013427}.

In this paper, we propose \emph{deep distributed recurrent Q-networks} (DDRQN) to enable teams of agents to learn effectively coordinated policies on such challenging problems.  We show that a naive approach to simply training independent DQN agents with \emph{long short-term memory} (LSTM) networks \cite{hochreiter1997long} is inadequate for multi-agent partially observable problems.    

Therefore, we introduce three modifications that are key to DDRQN's success: a) \emph{last-action inputs}: supplying each agent with its previous action as input on the next time step so that agents can approximate their action-observation histories; b) \emph{inter-agent weight sharing}: a single network's weights are used by all agents but that network conditions on the agent's unique ID, to enable fast learning while also allowing for diverse behaviour; and c) \emph{disabling experience replay}, which is poorly suited to the non-stationarity arising from multiple agents learning simultaneously.

To evaluate DDRQN, we propose two multi-agent reinforcement learning problems that are based on well-known riddles:
the \emph{hats riddle}, where $n$ prisoners in a line must determine their own hat colours; and the \emph{switch riddle}, in which $n$ prisoners must determine when they have all visited a room containing a single switch. Both riddles have been used as interview questions at companies like Google and Goldman Sachs.  

While these environments do not require convolutional networks for perception, the presence of partial observability means that they do require recurrent networks to deal with complex sequences, as in some single-agent works \cite{hausknecht2015deep,Ba:2015} and language-based \cite{Narasimhan:2015} tasks.  In addition, because partial observability is coupled with multiple agents, optimal policies critically rely on communication between agents. Since no communication protocol is given a priori, reinforcement learning must automatically develop a coordinated communication protocol.

Our results demonstrate that DDRQN can successfully solve these tasks, outperforming baseline methods, and discovering elegant communication protocols along the way. To our knowledge, this is the first time deep reinforcement learning has succeeded in learning communication protocols. In addition, we present ablation experiments that confirm that each of the main components of the DDRQN architecture are critical to its success.

\section{Background}
\label{sec:background}

In this section, we briefly introduce DQN and its multi-agent and recurrent extensions.

\subsection{Deep Q-Networks}
\label{subsec:dqn}

In a single-agent, fully-observable, reinforcement learning setting \cite{SuttonBarto:1998}, an agent observes its current state $s_t \in \mathcal{S}$ at each discrete time step $t$, chooses an action $a_t \in \mathcal{A}$ according to a potentially stochastic policy $\pi$, observes a reward signal $r_t$, and transitions to a new state $s_{t+1}$. Its objective is to maximize an expectation over the discounted return, $R_t$
\be
R_t =  r_t + \gamma r_{t+1} + \gamma^2 r_{t+2} + \cdots, 
\ee
where
$\gamma \in [0,1)$ is a discount factor.
The $Q$-function of a policy $\pi$ is:
\be
Q^{\pi}(s,a) = \E \left[  R_t | s_t = s, a_t = a \right].
\ee
The optimal action-value function $Q^{*}(s,a) = \max_{\pi} Q^{\pi}(s,a)$ obeys the Bellman optimality equation:
\be
Q^{*}(s,a) = \E_{s'} \left[ r +  \gamma  \max_{a' } Q^{*}(s',a')~|~s,a \right]. 
\ee
Deep $Q$-networks \cite{Mnih:2015} (DQNs) use neural networks parameterised by $\theta$ to represent $Q(s,a;\theta)$. DQNs are optimised by minimising the following loss function at each iteration~$i$:
\be
L_i(\theta_i) = \E_{s,a,r,s'} \left[ \left( y_i^{DQN} - Q(s,a; \theta_{i}) \right)^{2} \right],
\label{eq:loss}
\ee
with target
\be
y_i^{DQN} =  r + \gamma \max_{a'} Q(s',a';\theta_{i}^{-}).
\ee
Here, $\theta_{i}^{-}$ are the weights of a target network that is frozen for a number of iterations while updating the online network $Q(s,a; \theta_{i})$ by gradient descent. 
DQN uses \emph{experience replay}~\citep{Lin:1993,Mnih:2015}: during learning, the agent builds a dataset $\mathcal{D}_t =\{e_1, e_2, \ldots, e_t\}$ of experiences
$e_t=(s_t, a_t, r_t, s_{t+1})$ across episodes.  The $Q$-network is then trained by sampling mini-batches of experiences  from $\mathcal{D}$ uniformly at random.
Experience replay helps prevent divergence by breaking correlations among the samples. It also enables reuse of past experiences for learning, thereby reducing sample costs.

\subsection{Independent DQN}
\label{subsec:i-dqn}

DQN has been extended to cooperative multi-agent settings, in which each agent $m$ observes the global $s_t$, selects an individual action $a^m_t$, and receives a team reward, $r_t$, shared among all agents.  \citet{tampuu2015multiagent} address this setting with a framework that combines DQN with \emph{independent Q-learning}, applied to two-player pong, in which all agents independently and simultaneously learn their own Q-functions $Q^m(s,a^m;\theta^m_i)$.  While independent Q-learning can in principle lead to convergence problems (since one agent's learning makes the environment appear non-stationary to other agents), it has a strong empirical track record \cite{Shoham:2007,MASfoundations09,Zawadzki:2014}.

\subsection{Deep Recurrent Q-Networks}
\label{subsec:po-dqn}
Both DQN and independent DQN assume full observability, i.e., the agent receives $s_t$ as input. By contrast, in partially observable environments, $s_t$ is hidden and instead the agent receives only an observation $o_t$ that is correlated with $s_t$ but in general does not disambiguate it.

\citet{hausknecht2015deep} propose the \emph{deep recurrent Q-network} (DRQN) architecture to address single-agent, partially observable settings.  Instead of approximating $Q(s,a)$ with a feed-forward network, they approximate $Q(o,a)$ with a recurrent neural network that can maintain an internal state and aggregate observations over time. This can be modelled by adding an extra input $h_{t-1}$ that represents the hidden state of the network, yielding $Q(o_t,h_{t-1},a; \theta_{i})$.
Thus, DRQN outputs both $Q_t$, and $h_t$, at each time step.
DRQN was tested on a partially observable version of ALE in which a portion of the input screens were blanked out.

\subsection{Partially Observable Multi-Agent RL}
\label{subsec:part-obs}

In this work, we consider settings where there are both multiple agents and partial observability: each agent receives its own private $o^m_t$ at each time step and maintains an internal state $h^m_t$.  However, we assume that learning can occur in a centralised fashion, i.e., agents can share parameters, etc., during learning so long as the policies they learn condition only on their private histories.  In other words, we consider centralised learning of decentralised policies.

We are interested in such settings because it is only when multiple agents and partial observability coexist that agents have the incentive to communicate.  Because no communication protocol is given a priori, the agents must first automatically develop and agree upon such a protocol.  To our knowledge, no work on deep RL has considered such settings and no work has demonstrated that deep RL can successfully learn communication protocols.

\section{DDRQN}
\label{sec:methods}

The most straightforward approach to deep RL in partially observable multi-agent settings is to simply combine DRQN with independent Q-learning, in which case each agent's $Q$-network represents $Q^m(o^m_t,h^m_{t-1},a^m; \theta^m_{i})$, which conditions on that agent's individual hidden state as well as observation.  This approach, which we call the \emph{naive method}, performs poorly, as we show in Section \ref{sec:evaluation}.

Instead, we propose \emph{deep distributed recurrent Q-networks} (DDRQN), which makes three key modifications to the naive method.  The first, \emph{last-action input}, involves providing each agent with its previous action as input to the next time step. Since the agents employ stochastic policies for the sake of exploration, they should in general condition their actions on their action-observation histories, not just their observation histories. Feeding the last action as input allows the RNN to approximate action-observation histories.  

The second, \emph{inter-agent weight sharing}, involves tying the weights of all agents networks.  In effect, only one network is learned and used by all agents.  However, the agents can still behave differently because they receive different observations and thus evolve different hidden states.  In addition, each agent receives its own index $m$ as input, making it easier for agents to specialise.  Weight sharing dramatically reduces the number of parameters that must be learned, greatly speeding learning.

The third, \emph{disabling experience replay}, simply involves turning off this feature of DQN.  Although experience replay is helpful in single-agent settings, when multiple agents learn independently the environment appears non-stationary to each agent, rendering its own experience obsolete and possibly misleading.

Given these modifications, DDRQN learns a $Q$-function of the form $Q(o^m_t,h^m_{t-1}, m,a^m_{t-1},a^m_t; \theta_{i})$.  Note that $\theta_i$ does not condition on $m$, due to weight sharing, and that $a^m_{t-1}$ is a portion of the history while $a^m_t$ is the action whose value the $Q$-network estimates.

\begin{algorithm}[tb]
   \caption{DDRQN}
   \label{alg:ddrqn}
   \hskip -2em
   \begin{adjustwidth}{-0.5em}{}
\begin{algorithmic} 
\State Initialise $\theta_1$ and $\theta^-_1$
\For{each episode $e$}
    \State $h^m_1 = \mathbf 0$ for each agent $m$
    \State $s_1 = $ initial state, $t = 1$
    \While{$s_t \ne$ terminal {\bfseries and} $t < T$}
        \For{each agent $m$}
            \State With probability $\epsilon$ pick random  $a^m_t$
            \State else $a^m_t = \argmax_a Q(o^m_t,h^m_{t-1}, m,a^m_{t-1},a; \theta_{i})$ 
        \EndFor
        \State Get reward $r_t$ and next state $s_{t+1}$, $t = t + 1$
    \EndWhile
    \State $\nabla \theta = 0$ \Comment{reset gradient}
    \For{$j=t-1$ {\bfseries to} $1$, $-1$}
        \For{each agent $m$}
            \State  $y^m_j = $ \hspace{-0.2em} \Big\{ \hspace{-0.6em}\begin{tabular}{ll} $r_j$,  if $s_j$ terminal,  else \\  $r_j  + \gamma \max_{a} Q(o^m_{j+1},h^m_{j}, m,a^m_{j},a; \theta^-_{i}) $ \end{tabular}
            \State Accumulate gradients for:
            \State $(y^m_j - Q(o^m_j,h^m_{j-1}, m,a^m_{j-1},a^m_j; \theta_{i}))^2$
        \EndFor
    \EndFor
    \State  $\theta_{i+1} = \theta_i + \alpha \nabla \theta$ \Comment{update parameters}
    \State  $\theta^{-}_{i+1} =  \theta^-_i  +  \alpha^{-} ( \theta_{i+1} - \theta^-_i)$ $\triangleright$ update target network
\EndFor
\end{algorithmic}
\end{adjustwidth}
\end{algorithm}

Algorithm \ref{alg:ddrqn} describes DDRQN. First, we initialise the target and Q-networks. For each episode, we also initialise the state, $s_1$, the internal state of the agents, $h^m_1$, and $a^m_0$. Next, for each time step we pick an action for each agent $\epsilon$-greedily w.r.t.\ the Q-function. We feed in the previous action, $a^m_{t-1}$, the agent index, $m$, along with the observation $o^m_t$ and the previous internal state, $h^m_{t-1}$. After all agents have taken their action, we query the environment for a state update and reward information. 

When we reach the final time step or a terminal state, we proceed to the Bellman updates. Here, for each agent, $m$, and time step, $j$, we calculate a target Q-value, $y^m_j$, using the observed reward, $r_j$, and the discounted target network. We also accumulate the gradients, $\nabla \theta$,  by regressing the Q-value estimate, $Q(o^m_j,h^m_{j-1}, m,a^m_{j-1},a; \theta_{i})$, against the target Q-value, $y^m_j$, for the action chosen, $a^m_j$.

Lastly, we conduct two weight updates, first $\theta_i$ in the direction of the accumulated gradients, $\nabla \theta$,  and then the target network, $\theta^-_i$, in the direction of $\theta_i$.

\section{Multi-Agent Riddles}
\label{sec:multiagent}

In this section, we describe the riddles on which we evaluate DDRQN.

\subsection{Hats Riddle}
\label{subsec:hats_riddle}

\begin{figure}[h]
    \centering
    \includegraphics[width=1\linewidth]{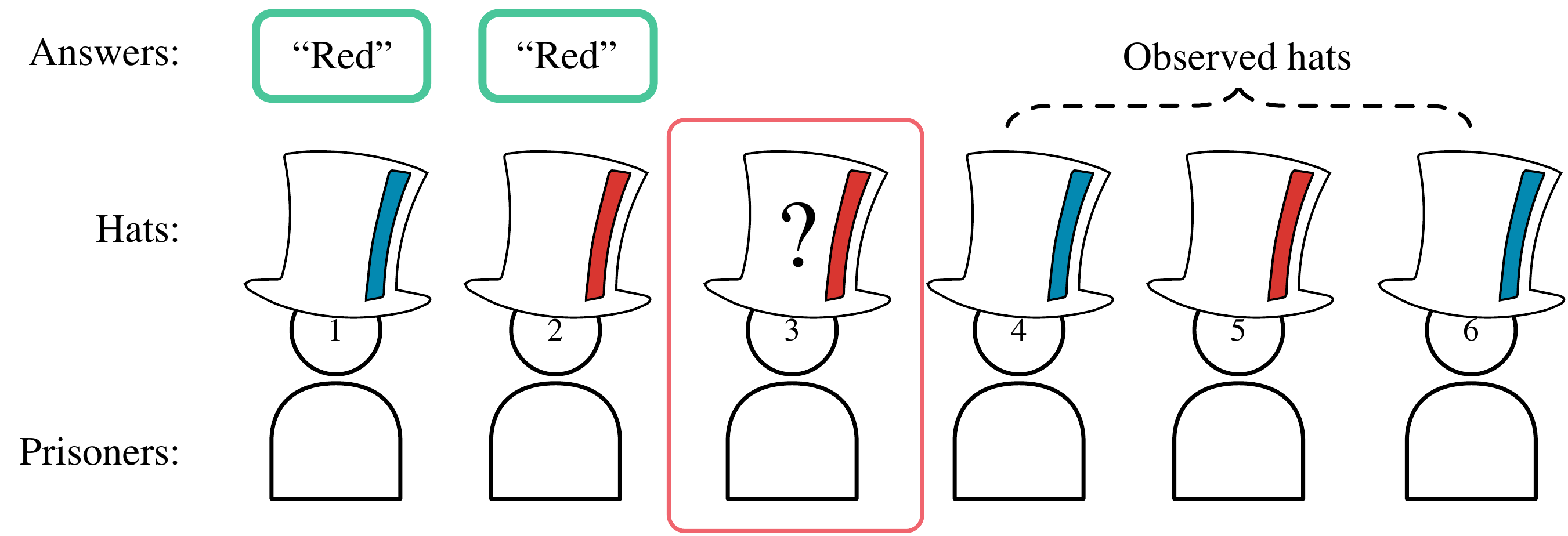} 
    \vskip -0.5em
    \caption{\emph{Hats:} Each prisoner can hear the answers from all preceding prisoners (to the left) and see the colour of the hats in front of him (to the right) but must guess his own hat colour.}
    \label{fig:hats_vis}
\end{figure}

The hats riddle can be described as follows:
\emph{``An executioner lines up 100 prisoners single file and puts a red or a blue hat on each prisoner's head. Every prisoner can see the hats of the people in front of him in the line - but not his own hat, nor those of anyone behind him. The executioner starts at the end (back) and asks the last prisoner the colour of his hat. He must answer ``red'' or ``blue.'' If he answers correctly, he is allowed to live. If he gives the wrong answer, he is killed instantly and silently. (While everyone hears the answer, no one knows whether an answer was right.) On the night before the line-up, the prisoners confer on a strategy to help them. What should they do?''} \cite{poundstone2012you}. Figure~\ref{fig:hats_vis} illustrates this setup.

An optimal strategy is for all prisoners to agree on a communication protocol in which the first prisoner says ``blue'' if the number of blue hats is even and ``red" otherwise (or vice-versa).  All remaining prisoners can then deduce their hat colour given the hats they see in front of them and the responses they have heard behind them.  Thus, everyone except the first prisoner will definitely answer correctly. 

To formalise the hats riddle as a multi-agent RL task, we define a state space $\mathbf{s} = (s^1,\ldots,s^n,a^1,\ldots,a^{n})$, where $n$ is the total number of agents, $s^m \in \{\text{blue, red}\}$ is the $m$-th agent's hat colour and $a^m \in \{\text{blue, red}\}$ is the action it took on the $m$-th step.  At all other time steps, agent $m$ can only take a null action. On the $m$-th time step, agent $m$'s observation is ${o^m = (a^1,\ldots,a^{m-1},s^{m+1},\ldots,s^n)}$. Reward is zero except at the end of the episode, when it is the total number of agents with the correct action: $r_{n} =\sum_m \mathds{I}( a^m = s^m )$. We label only the relevant observation $o^m$ and action $a^m$ of agent $m$, omitting the time index.  

Although this riddle is a single action and observation problem it is still partially observable, given that none of the agents can observe the colour of their own hat.

\subsection{Switch Riddle}
\label{subsec:switch-riddle}

The switch riddle can be described as follows:
\emph{``One hundred prisoners have been newly ushered into prison. The warden tells them that starting tomorrow, each of them will be placed in an isolated cell, unable to communicate amongst each other. Each day, the warden will choose one of the prisoners uniformly at random with replacement, and place him in a central interrogation room containing only a light bulb with a toggle switch. The prisoner will be able to observe the current state of the light bulb. If he wishes, he can toggle the light bulb. He also has the option of announcing that he believes all prisoners have visited the interrogation room at some point in time. If this announcement is true, then all prisoners are set free, but if it is false, all prisoners are executed. The warden leaves and the prisoners huddle together to discuss their fate. Can they agree on a protocol that will guarantee their freedom?''}~\cite{wu2002100}. 

\begin{figure}[h]
    \centering
    \includegraphics[width=1\linewidth]{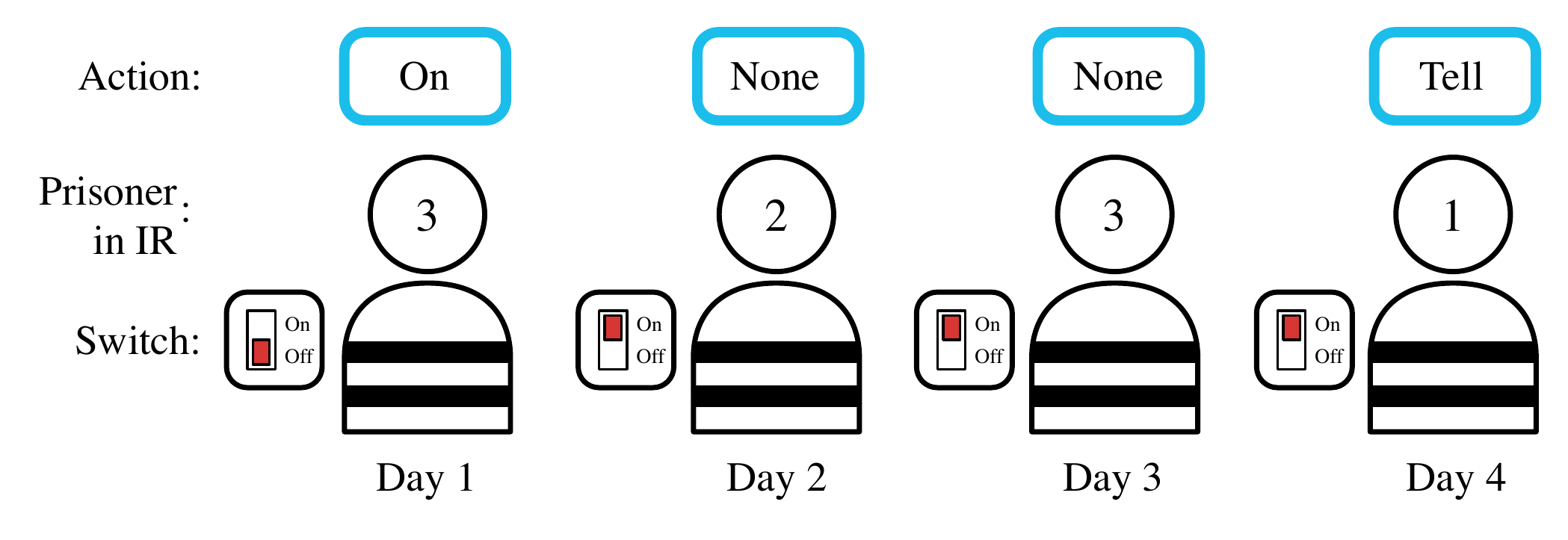}    
    \vskip -0.5em
    \caption{\emph{Switch:} Every day one prisoner gets sent to the interrogation room where he can see the switch and choose between actions ``On'', ``Off'', ``Tell'' and ``None''. } 
    \label{fig:switch_vis}
    \end{figure}

A number of strategies ~\cite{yisong2012100, wu2002100} have been analysed for the infinite time-horizon version of this problem in which the goal is to guarantee survival. One well-known strategy is for one prisoner to be designated the counter. Only he is allowed to turn the switch off while each other prisoner turns it on only once. Thus, when the counter has turned the switch off $n - 1$ times, he can ``Tell''. 

To formalise the switch riddle, we define a state space $\mathbf{s} = (SW_t, IR_t,s^1,\ldots,s^n)$, where $SW_t  \in \{\text{on,  off}\}$ is the position of the switch, $IR \in \{1 \ldots n\}$ is the current visitor in the interrogation room and $s^1,\ldots,s^n \in \{0,1\}$ tracks which agents have already been to the interrogation room.
At time step $t$, agent $m$ observes ${o^m_t = (ir_t, sw_t)}$, where ${ir_t = \mathds{I}(IR_t = m)}$, and  $sw_t = SW_t$ if the agent is in the interrogation room and null otherwise. If agent $m$ is in the interrogation room then its actions are $a^m_t \in \{\text{``On'', ``Off'', ``Tell'', ``None''}\}$; otherwise the only action is "None". The episode ends when an agent chooses ``Tell'' or when the maximum time step is reached. The reward $r_t$ is 0 except unless an agent chooses ``Tell'', in which case it is 1 if all agents have been to the interrogation room and $-1$ otherwise.

\section{Experiments}
\label{sec:evaluation}

In this section, we evaluate DDRQN on both multi-agent riddles. In our experiments, prisoners select actions using an $\epsilon$-greedy policy with $\epsilon = 1 - 0.5^\frac{1}{n}$ for the hats riddle and $\epsilon=0.05$ for the switch riddle. For the latter, the discount factor was set to $\gamma=0.95$, and the target networks, as described in Section~\ref{sec:methods}, update with  $\alpha^{-}=0.01$, while in both cases weights were optimised using Adam \cite{kingma2014adam} with a learning rate of \num{1e-3}. The proposed architectures make use of rectified linear units, and LSTM cells. Further details of the network implementations are described in the Supplementary Material and source code will be published online.

\subsection{Hats Riddle}
\label{subsec:exp_hats}

\begin{figure}[tb]
    \centering
    \includegraphics[width=1\linewidth]{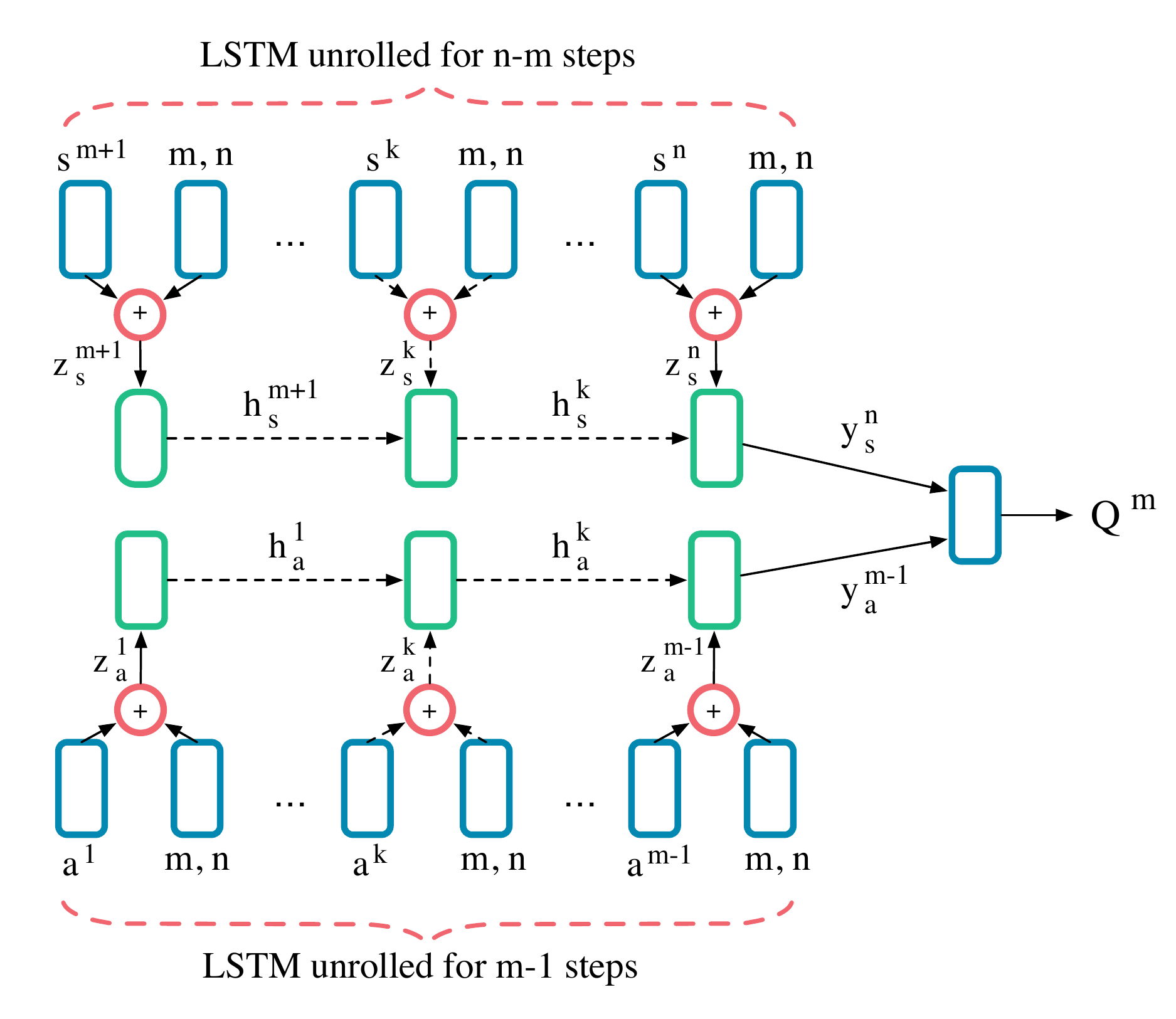}
    \figspace
    \caption{\emph{Hats:} Each agent $m$ observes the answers, $a^k$, $k<m$, from all preceding agents and hat colour, $s^k$ in front of him, $k > m$.  Both variable length sequences are processed through RNNs. First, the answers heard are passed through two single-layer MLPs, $z^k_a =\MLP{a^k} \oplus \MLP{m,n}$, and their outputs are added element-wise. $z^k_a$ is passed through an LSTM network $y^{k}_a, h^{k}_a = \text{LSTM}_a(z^k_a, h^{k-1}_a)$. Similarly for the observed hats we define $y^{k-1}_s,h^{k-1}_s = \text{LSTM}_s(z^k_s, h^{k-1}_s)$.
The last values of the two LSTMs $y^{m-1}_a$ and $y^{n}_s$ are used to approximate $Q^m = \text{MLP}(y^{m-1}_a||y^{n}_s)$ from which the action $a^m$ is chosen.}
    \label{fig:hats_arch}
\end{figure}

Figure~\ref{fig:hats_arch} shows the architecture we use to apply DDRQN to the hats riddle.
To select $a^m$, the network is fed as input $o^m = (a^1,\ldots,a^{m-1},s^{m+1},\ldots,s^n)$, as well as $m$ and $n$.
The answers heard are passed through two single-layer MLPs, $z^k_a=\text{MLP}\lbrack 1\times64\rbrack (a^k) \oplus \text{MLP}\lbrack 2\times64\rbrack (m,n)$, and their outputs are added element-wise. Subsequently, $z^k_a$ is passed through an LSTM network $y^k_a,h^k_a = \text{LSTM}_a \lbrack 64\rbrack (z^k_a, h^{k-1}_a)$. We follow a similar procedure for the $n-m$ hats observed defining $y^k_s,h^k_s = \text{LSTM}_s \lbrack 64\rbrack (z^k_s, h^{k-1}_s)$.
Finally, the last values of the two LSTM networks $y^{m-1}_a$ and $y^n_s$ are used to approximate the Q-Values for each action $Q^m = \text{MLP} \lbrack 128\times64,64\times64,64\times1\rbrack (y^{m-1}_a||y^n_s)$. The network is trained with mini-batches of $20$ episodes.

\begin{figure}[b!]
    \vskip -0.3in
    \centering
    \includegraphics[width=1\linewidth]{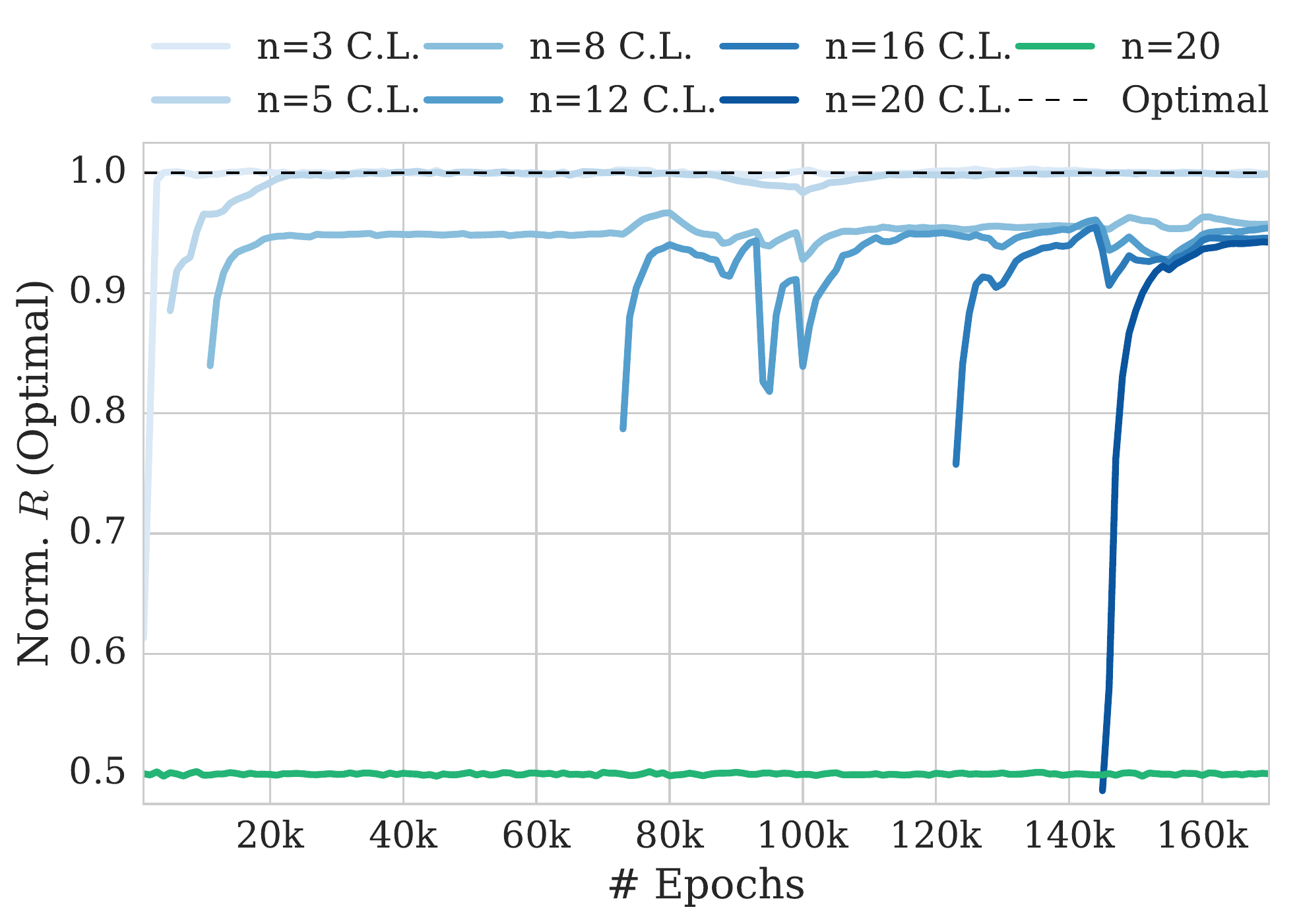}
    \figspace
    \caption{\emph{Hats}: Using Curriculum Learning DDRQN achieves good performance for $n = 3 . . . 20$ agents, compared to the optimal strategy.}
    \label{fig:h_20_curriculum}
\end{figure}

Furthermore, we use an adaptive variant of \emph{curriculum learning}~\cite{bengio2009curriculum} to pave the way for scalable strategies and better training performance. We sample examples from a multinomial distribution of curricula, each corresponding to a different $n$, where the current bound is raised every time performance becomes near optimal. 
The probability of sampling a given $n$ is inversely proportional to the performance gap compared to the normalised maximum reward. The performance is depicted in Figure~\ref{fig:h_20_curriculum}.

\begin{figure}[!b]
    \centering
    \includegraphics[width=1\linewidth]{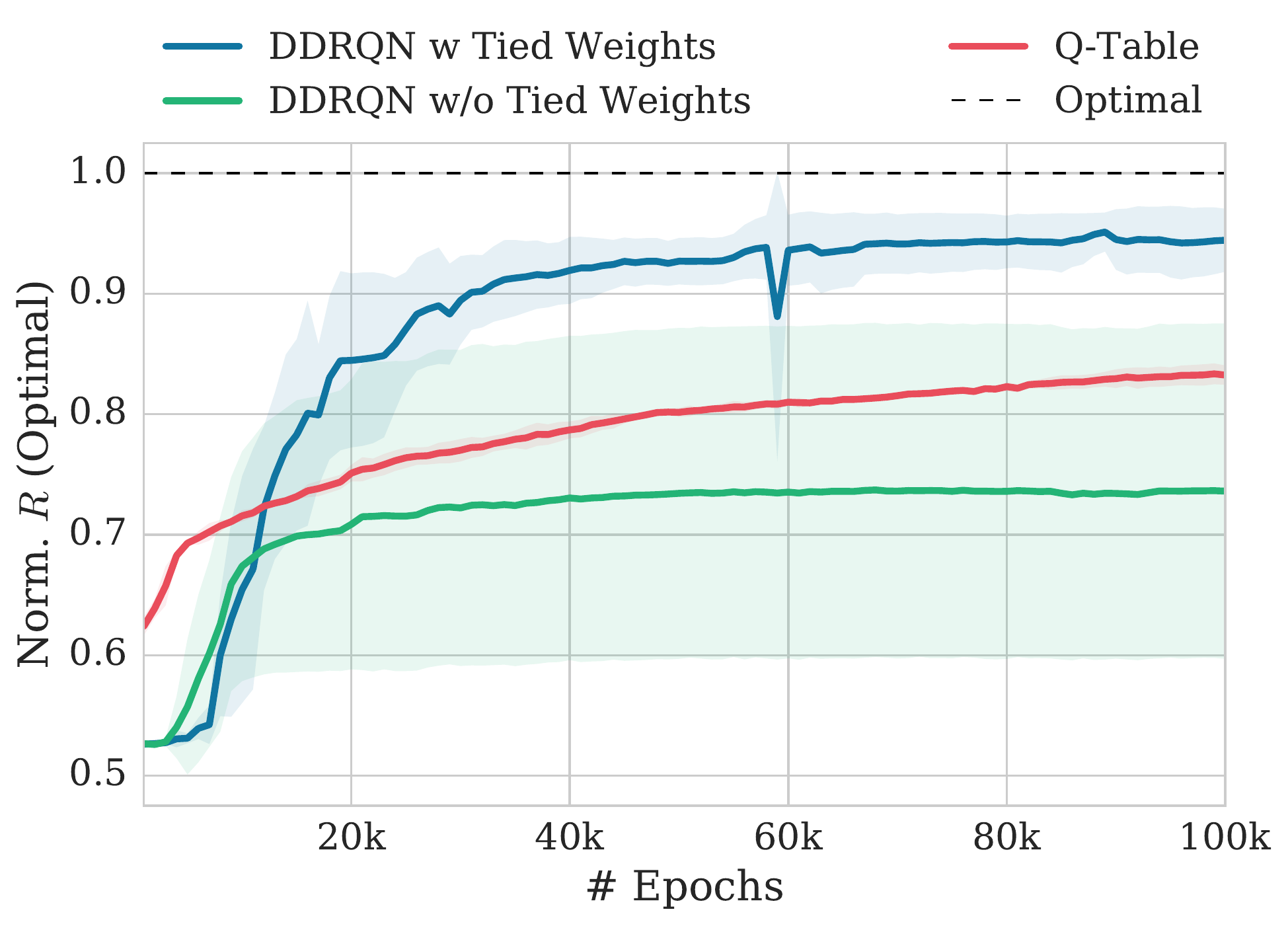}
    \figspace
    \caption{Results on the hats riddle with $n = 10$ agents, comparing DDRQN with and without inter-agent weight sharing to a tabular Q-table and a hand-coded optimal strategy. The lines depict the average of $10$ runs and $95\%$ confidence intervals.}
    \label{fig:h_10_shared}
\end{figure}

We first evaluate DDRQN for $n = 10$ and compare it with tabular Q-learning. Tabular Q-learning is feasible only with few agents, since the state space grows exponentially with $n$.  In addition, separate tables for each agent precludes generalising across agents.

Figure~\ref{fig:h_10_shared} shows the results, in which DDRQN substantially outperforms tabular Q-learning.  In addition, DDRQN also comes near in performance to the optimal strategy described in Section \ref{subsec:hats_riddle}. This figure also shows the results of an ablation experiment in which inter-agent weight sharing has been removed from DDRQN.  The results confirm that inter-agent weight sharing is key to performance.  

Since each agent takes only one action in the hats riddle, it is essentially a single step problem.  Therefore, last-action inputs and disabling experience replay do not play a role and do not need to be ablated.  We consider these components in the switch riddle in Section~\ref{subsec:exp_switches}.

\begin{table}[t]
\caption{Percent agreement on the hats riddle between DDRQN and the optimal parity-encoding strategy.}
\label{table:hats-strategy}
\vskip 0.15in
\begin{center}
\begin{small}
\begin{sc}
\begin{tabular}{cc}
\hline
\abovespace\belowspace
n & \% Agreement \\
\hline
\abovespace
3  & $100.0\%$ \\
5  & $100.0\%$ \\
8  & $79.6\%$ \\
12 & $52.6\%$ \\
16 & $50.8\%$ \\
20 & $52.5\%$ \\
\hline
\end{tabular}
\end{sc}
\end{small}
\end{center}
\vskip -0.2in
\end{table}

We compare the strategies DDRQN learns to the optimal strategy by computing the percentage of trials in which the first agent correctly encodes the parity of the observed hats in its answer. Table~\ref{table:hats-strategy} shows that the encoding is almost perfect for $n \in \{3, 5, 8\}$. For $n \in \{12,16,20\}$, the agents do not encode parity but learn a different distributed solution that is nonetheless close to optimal. We believe that qualitatively this solution corresponds to more the agents communicating information about other hats through their answers, instead of only the first agent.

\subsection{Switch Riddle}
\label{subsec:exp_switches}

\begin{figure}[!t]
    \centering
    \includegraphics[width=1\linewidth]{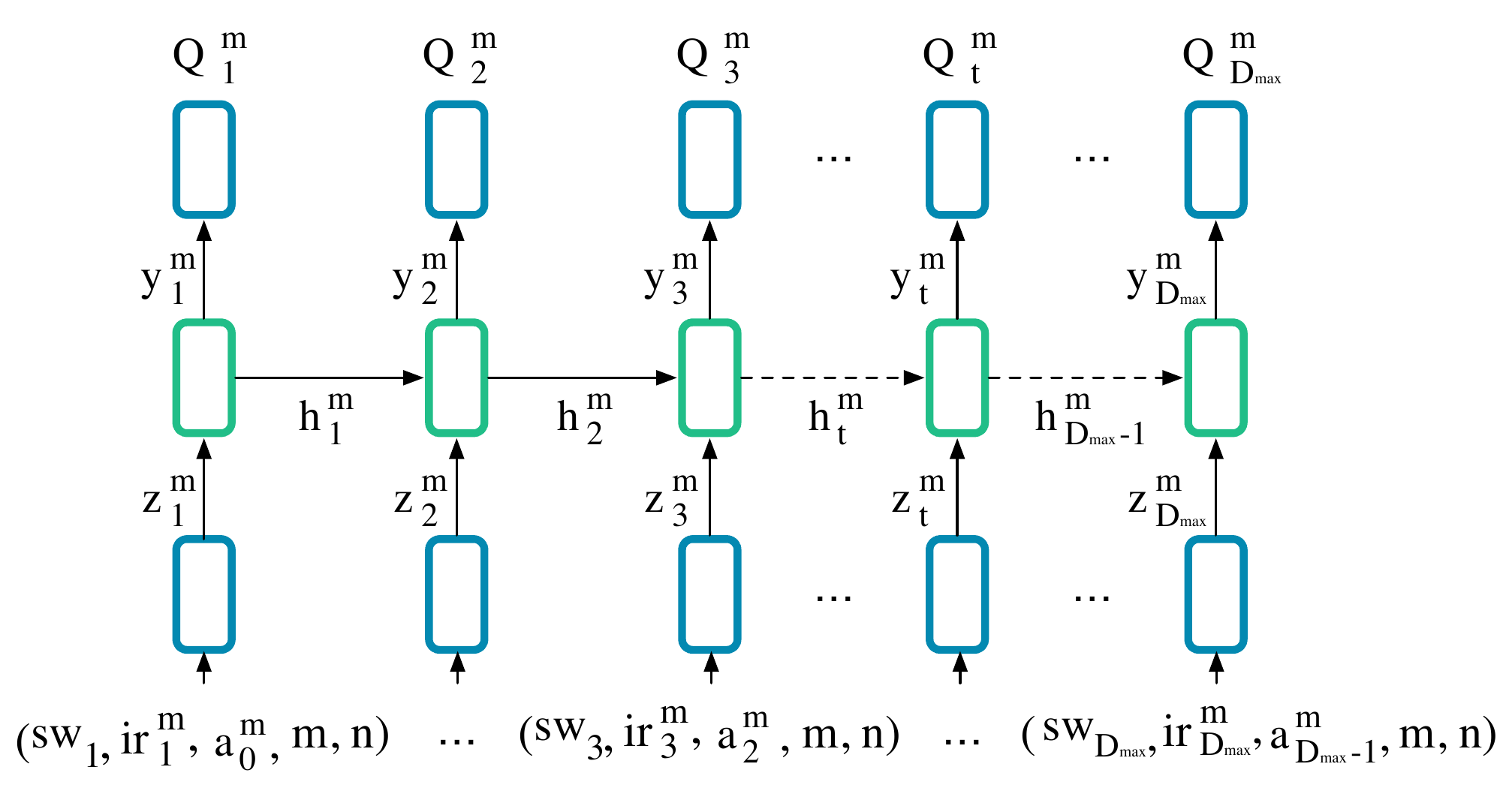}
    \figspace
    \caption{\emph{Switch:} Agent $m$ receives as input: the switch state $sw_t$, the room he is in, $ir^m_t$,  his last action, $a^m_{t-1}$, his ID, $m$, and the $\#$ of agents, $n$. At each step, the inputs are processed through a 2-layer MLP $z^m_t = \text{MLP} (sw_t, ir^m_t, \text{OneHot}(a^m_{t-1}), \text{OneHot}(m), n)$. Their embedding $z^m_t$ is then passed to an LSTM network, $y^m_t, h^m_t = \text{LSTM} (z^m_t, h^m_{t-1})$, which is used to approximate the agent's action-observation history. Finally, the output $y^m_t$ of the LSTM is used at each step to compute $Q^m_t = \text{MLP}(y^m_t)$.}
    \label{fig:switch_arch}
    \figspace
\end{figure}

Figure~\ref{fig:switch_arch} illustrates the model architecture used in the switch riddle. Each agent $m$ is modelled as a recurrent neural network with LSTM cells that is unrolled for $D_{\max}$ time-steps, where $d$ denotes the number of days of the episode. In our experiments, we limit $d$ to $D_{\max} = 4n-6$ in order to keep the experiments computationally tractable.

\begin{figure}[tb]
    \centering
    \includegraphics[width=1\linewidth]{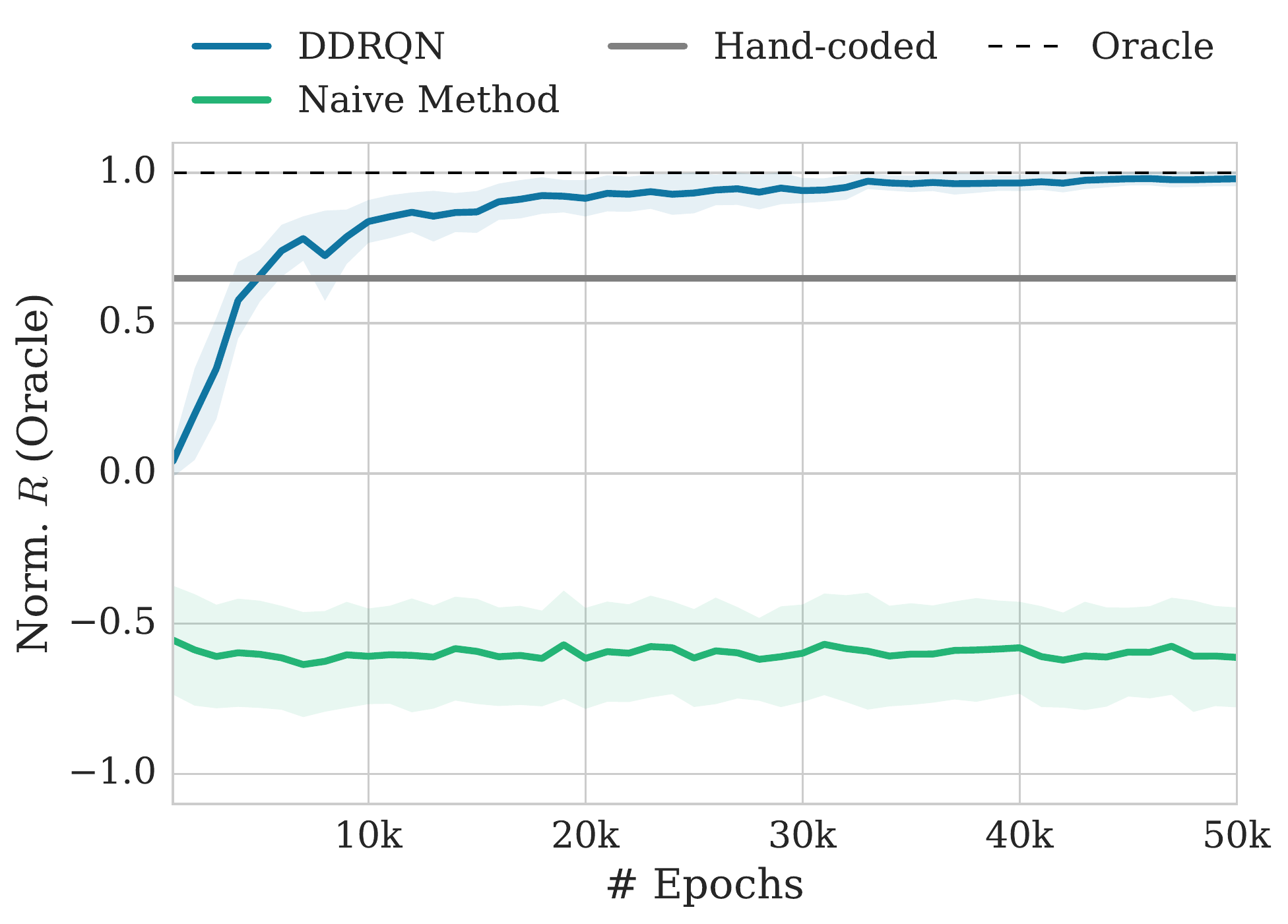}
    \figspace
    \caption{\emph{Switch:} For $n = 3$ DDRQN outperforms the ``Naive Method'' and a simple hand coded strategy, ``tell on last day'', achieving ``Oracle'' level performance. The lines depict the average of $10$ runs and the $95\%$ confidence interval.}
    \label{fig:s_3_3_vs}
\end{figure}

The inputs, $o^m_t, a^m_{t-1}, m$ and $n$,  are processed through a 2-layer MLP $z^m_t = \text{MLP}\lbrack (7+n)\times128, 128\times128\rbrack (o^m_t, \text{OneHot}(a^m_{t-1}), \text{OneHot}(m), n)$. Their embedding $z^m_t$ is then passed an LSTM network, $y^m_t, h^m_t = \text{LSTM}\lbrack 128 \rbrack (z^m_t, h^m_{t-1})$, which is used to approximate the agent's action-observation history. Finally, the output $y^m_t$ of the LSTM is used at each step to approximate the Q-values of each action using a 2-layer MLP $Q^m_t = \text{MLP}\lbrack 128\times128, 128\times128, 128\times4 \rbrack (y^m_t)$. 
As in the hats riddle, curriculum learning was used for training.

Figure~\ref{fig:s_3_3_vs}, which shows results for $n = 3$, shows that DDRQN learns an optimal policy, beating the naive method and the hand coded strategy, ``tell on last day''. This verifies that the three modifications of DDRQN substantially improve performance on this task. In following paragraphs we analyse the importance of the individual modifications.

\begin{figure}[!b]
    \centering
    \includegraphics[width=0.8\linewidth]{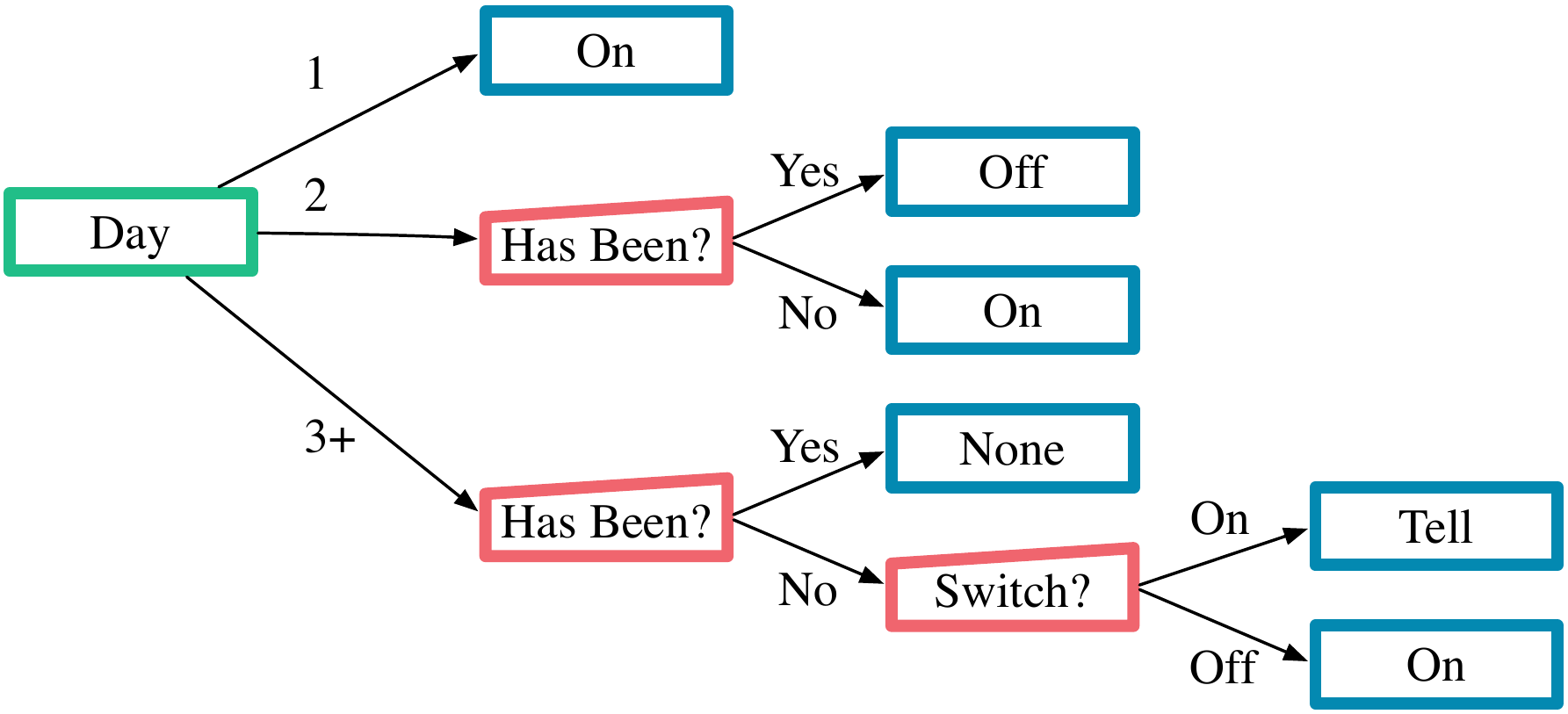}
    \figspace
    \caption{\emph{Switch:} For $n=3$, DDRQN manages to discover a perfect strategy, which we visualise as a decision tree in this Figure. After day $2$, the ``on'' position of the switch encodes that 2 prisoners have visited the interrogation room, while ``off'' encodes that one prisoner has.}
    \label{fig:switch_strategy}
\end{figure}

We analysed the strategy that DDRQN discovered for $n=3$ by looking at 1000 sampled episodes. Figure~\ref{fig:switch_strategy} shows a decision tree, constructed from those samples, that corresponds to an optimal strategy  allowing the agents to collectively track the number of visitors to the interrogation room. 
When a prisoner visits the interrogation room after day two, there are only two options: either one or two prisoners may have visited the room before. If three prisoners had been, the third prisoner would have already finished the game. The two remaining options can be encoded via the ``on'' and ``off'' position respectively.
In order to carry out this strategy each prisoner has to learn to keep track of whether he has visited the cell before and what day it currently is.

\begin{figure}[h]
    \centering
    \includegraphics[width=1\linewidth]{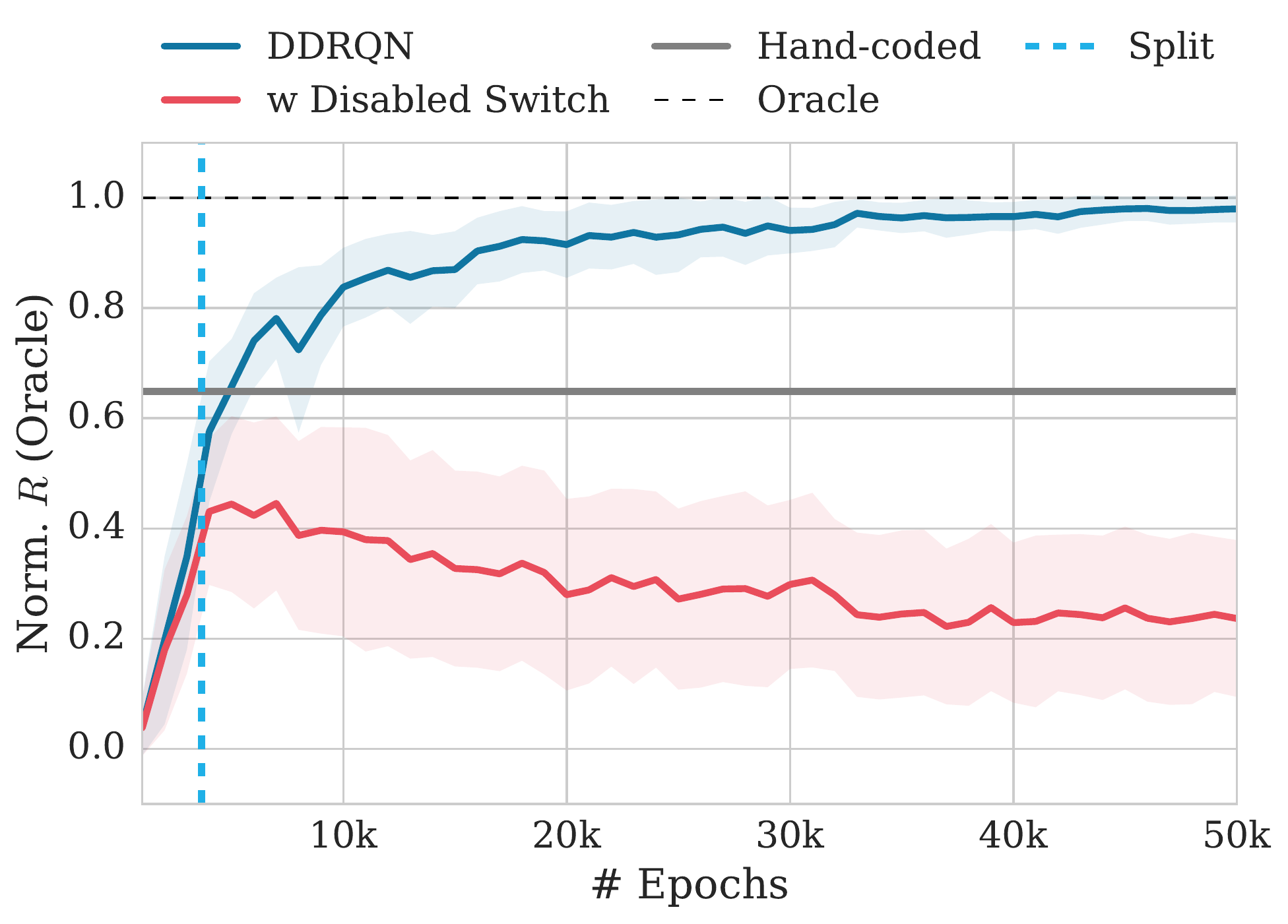}
    \figspace
    \caption{\emph{Switch:} At $3.5$k episodes the DDRQN line clearly separates from the performance line for the ``no switch'' test and start exceeding ``tell on last day''. At this point the agents start to discover strategies that evolve communication using the switch. The lines depict the mean of $10$ runs and the $95\%$ confidence interval.}
    \label{fig:s_3_3_noswitches}
\end{figure}

Figure~\ref{fig:s_3_3_noswitches} compares the performance of DDRQN to a variant in which the switch has been disabled.  After around 3,500 episodes, the two diverge in performance.  Hence, there is a clearly identifiable point during learning when the prisoners start learning to communicate via the switch.  Note that only when the switch is enabled can DDRQN outperform the hand-coded ``tell on last day'' strategy. Thus, communication via the switch is required for good performance.

\begin{figure}[!b]
    \centering
    \includegraphics[width=1\linewidth]{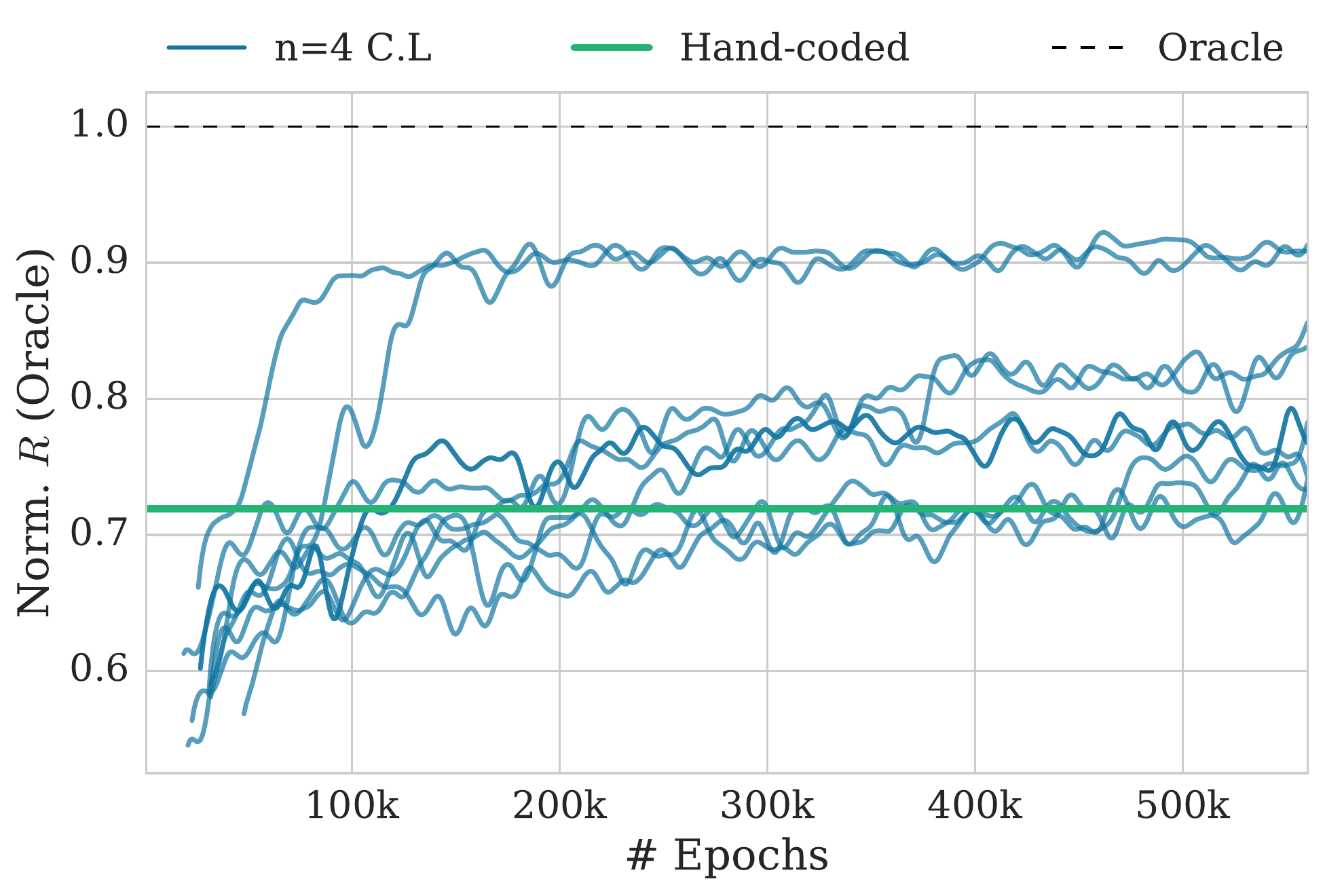}
    \figspace
    \caption{\emph{Switch:} $10$ runs using curriculum learning for $n=3,4$. In most cases was able DDRQN to find strategies that outperform ``tell on last day'' for $n = 4$.}
    \label{fig:s_3_3_cur}
\end{figure}
Figure~\ref{fig:s_3_3_cur} shows performance for $n = 4$. On most runs, DDRQN clearly beats the hand-coded ``tell on last day'' strategy and final performance approaches $90\%$ of the oracle. However, on some of the remaining runs DDRQN fails to significantly outperform the hand-coded strategy.
Analysing the learned strategies suggests that prisoners typically encode whether $2$ or $3$ prisoners have been to the room via the ``on'' and ``off'' positions of the switch, respectively. This strategy generates no false negatives, i.e., when the $4th$ prisoner enters the room, he always ``Tells'', but generates false positives around $5\%$ of the time.  Example strategies are included in the Supplementary Material.

\begin{figure}[h]
    \centering
    \includegraphics[width=1\linewidth]{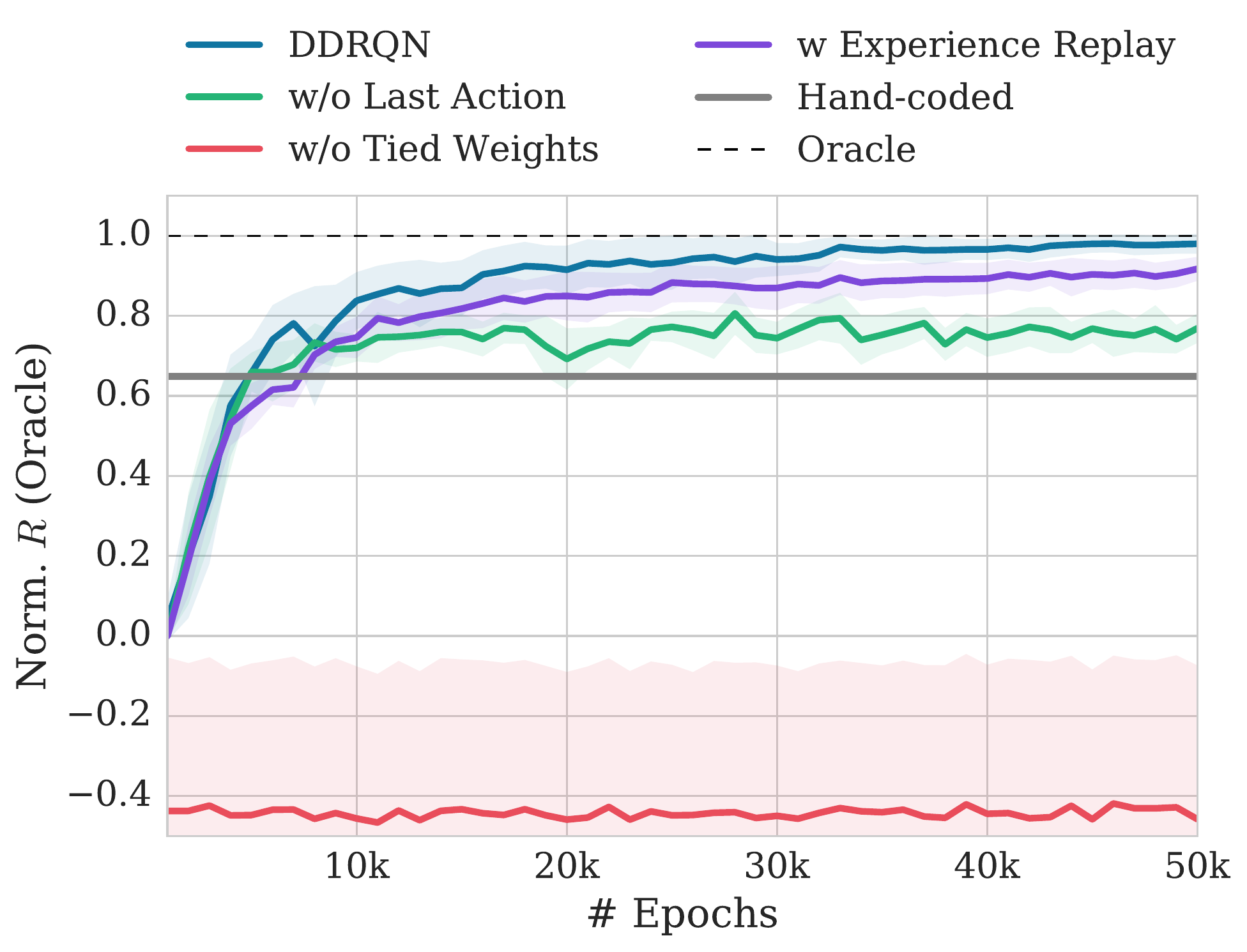}
    \figspace
    \caption{\emph{Switch:} Tied weights and last action input are key for the performance of DDRQN. Experience replay prevents agents from reaching Oracle level. The experiment was executed for $n = 3$ and the lines depict the average of $10$ runs and the $95\%$ confidence interval.}
    \label{fig:s_3_3_contributions}
\end{figure}

Furthermore, Figure~\ref{fig:s_3_3_contributions} shows the results of ablation experiments in which each of the modifications in DDRQN is removed one by one.  The results show that all three modifications contribute substantially to DDRQN's performance. Inter-agent weight sharing is by far the most important, without which the agents are essentially unable to learn the task, even for $n = 3$.   Last-action inputs also play a significant role, without which performance does not substantially exceed that of the ``tell on last day'' strategy.  Disabling experience replay also makes a difference, as performance with replay never reaches optimal, even after 50,000 episodes.  This result is not surprising given the non-stationarity induced by multiple agents learning in parallel. Such non-stationarity arises even though agents can track their action-observation histories via RNNs within a given episode. Since their memories are reset between episodes, learning performed by other agents appears as non-stationary from their perspective.

However, it is particularly important in communication-based tasks like these riddles, since the value function for communication actions depends heavily on the interpretation of these messages by the other agents, which is in turn set by their Q-functions.

\section{Related Work}
\label{sec:related_work}
There has been a plethora of work on multi-agent reinforcement learning with communication, e.g., \cite{tan1993multi,melo2011querypomdp,Panait2005387,Zhang20131101,Maravall2013661}. However, most of this work assumes a pre-defined communication protocol. One exception is the work of \citet{kasai2008learning}, in which the tabular Q-learning agents have to learn the content of a message to solve a predator-prey task. Their approach is similar to the Q-table benchmark used in Section~\ref{subsec:exp_hats}. By contrast, DDRQN uses recurrent neural networks that allow  for memory-based communication and generalisation across agents. 

Another example of open-ended communication learning in a multi-agent task is given in \cite{giles2002learning}. However, here evolutionary methods are used for learning communication protocols, rather than RL. By using deep RL with shared weights we enable our agents to develop distributed communication strategies and to allow for faster learning via gradient based optimisation. 
    
Furthermore, planning-based RL methods have been employed to include messages as an integral part of the multi-agent reinforcement learning challenge  \cite{spaan2006decentralized}. However, so far this work has not been extended to deal with high dimensional complex problems. 

In ALE, partial observability has been artificially introduced by blanking out a fraction of the input screen~\cite{hausknecht2015deep}. Deep recurrent reinforcement learning has also been applied to text-based games, which are naturally partially observable \cite{Narasimhan:2015}. Recurrent DQN was also successful in the email campaign challenge \cite{Li:2015}.

However, all these examples apply recurrent DQN in single-agent domains. Without the combination of multiple agents and partial observability, there is no need to learn a communication protocol, an essential feature of our work.

\section{Conclusions \& Future Work}

This paper proposed \emph{deep distributed recurrent Q-networks} (DDRQN), which enable teams of agents to learn to solve communication-based coordination tasks.  In order to successfully communicate, agents in these tasks must first automatically develop and agree upon their own communication protocol.  We presented empirical results on two multi-agent learning problems based on well-known riddles, demonstrating that DDRQN can successfully solve such tasks and discover elegant communication protocols to do so.  In addition, we presented ablation experiments that confirm that each of the main components of the DDRQN architecture are critical to its success.

Future work is needed to fully understand and improve  the scalability of the DDRQN architecture for large numbers of agents, e.g., for $n > 4$  in the switch riddle. We also
hope to further explore the ``local minima'' structure of the coordination and strategy space that underlies these riddles. Another avenue for improvement is to extend DDRQN to make use of various multi-agent adaptations of Q-learning \cite{tan1993multi, littman1994markov, lauer2000algorithm, Panait2005387}. 

A benefit of using deep models is that they can efficiently cope with high dimensional perceptual signals as inputs. In the future this can be tested by replacing the binary representation of the colour with real images of hats or applying DDRQN to other scenarios that involve real world data as input.

While we have advanced a new proposal for using riddles as a test field for multi-agent partially observable reinforcement learning with communication, we also hope that this research will spur the development of further interesting and challenging domains in the area.

\section{Acknowledgements}
This work was supported by the Oxford-Google DeepMind Graduate Scholarship and the EPSRC.
%


\bibliography{refs,deeprl}
\bibliographystyle{include/icml2016}



\end{document}